\documentclass[conference]{IEEEtran}
\ifCLASSINFOpdf
   \usepackage[pdftex]{graphicx}
   \usepackage[skip=2pt, font=footnotesize, labelsep=period]{caption}
   \usepackage{multirow}
   \usepackage{tabularew}
   \usepackage[flushleft]{threeparttable}

\usepackage{subfig}

\else
\fi
\IEEEoverridecommandlockouts

\begin{document}
%
\title{Significance Driven Hybrid 8T-6T SRAM for Energy-Efficient Synaptic Storage in Artificial Neural Networks}

\author{\IEEEauthorblockN{ Gopalakrishnan Srinivasan, Parami Wijesinghe, Syed Shakib Sarwar, Akhilesh Jaiswal, and Kaushik Roy}
\IEEEauthorblockA{School of Electrical and Computer Engineering,
Purdue University\\
Email: \{srinivg, pwijesin, sarwar, jaiswal, kaushik\}@purdue.edu}
\thanks {This work was supported in part by C-SPIN, one of the six centers of STARnet, a Semiconductor Research Corporation program, sponsored by MARCO and DARPA, by the Semiconductor Research Corporation, the National Science Foundation, and Intel Corporation.}
\\[-3.0ex]
}


%


\maketitle

\begin{abstract}
Multilayered artificial neural networks have found widespread utility in classification and recognition applications. The scale and complexity of such networks together with the inadequacies of general purpose computing platforms have led to a significant interest in the development of efficient hardware implementations. In this work, we focus on designing energy-efficient on-chip storage for the synaptic weights, motivated primarily by the observation that the number of synapses is orders of magnitude larger than the number of neurons. Typical digital CMOS implementations of such large-scale networks are power hungry. In order to minimize the power consumption, the digital neurons could be operated reliably at scaled voltages by reducing the clock frequency. On the contrary, the on-chip synaptic storage designed using a conventional 6T SRAM is susceptible to bitcell failures at reduced voltages. 
However, the intrinsic error resiliency of neural networks to small synaptic weight perturbations enables us to scale the operating voltage of the 6T SRAM. Our analysis on a widely used digit recognition dataset indicates that the voltage can be scaled by 200 mV from the nominal operating voltage (950 mV) for practically no loss (less than 0.5\%) in accuracy (22 nm predictive technology). Scaling beyond that causes substantial performance degradation owing to increased probability of failures in the MSBs of the synaptic weights. We, therefore propose a significance driven hybrid 8T-6T SRAM, wherein the sensitive MSBs are stored in 8T bitcells that are robust at scaled voltages due to decoupled read and write paths. In an effort to further minimize the area penalty, we present a synaptic-sensitivity driven hybrid memory architecture consisting of multiple 8T-6T SRAM banks.
Our circuit to system-level simulation framework shows that the proposed synaptic-sensitivity driven architecture provides a 30.91\% reduction in the memory access power with a 10.41\% area overhead, for less than 1\% loss in the classification accuracy. 
\end{abstract}
\setlength{\textfloatsep}{5pt}

%
\IEEEpeerreviewmaketitle


\section{Introduction}
The advancements in semiconductor technology have led to renewed interest in efficient implementations of complex neuromorphic systems. Artificial Neural Networks (ANNs) that consist of fully connected layers of neurons, as illustrated in Fig. 1, have been widely used in classification and recognition applications. It is primarily due to their inherent ability to learn sophisticated nonlinear mapping between large unstructured input and output data [1, 2]. Deep architectures, essentially inspired by the hierarchical organization of the human brain, are the cornerstone of contemporary neuromorphic systems. The superior performance of such multilayered neural networks can be attributed to hierarchical feature extraction, wherein the low level features uncovered by the initial layers are subsequently used to discern abstract patterns [3]. This has resulted in their widespread utility in a diverse suite of applications, including visual object classification [4], speech recognition [5], and dimensionality reduction [6].

\begin{figure}[!t]
\centering
\includegraphics[width=3in]{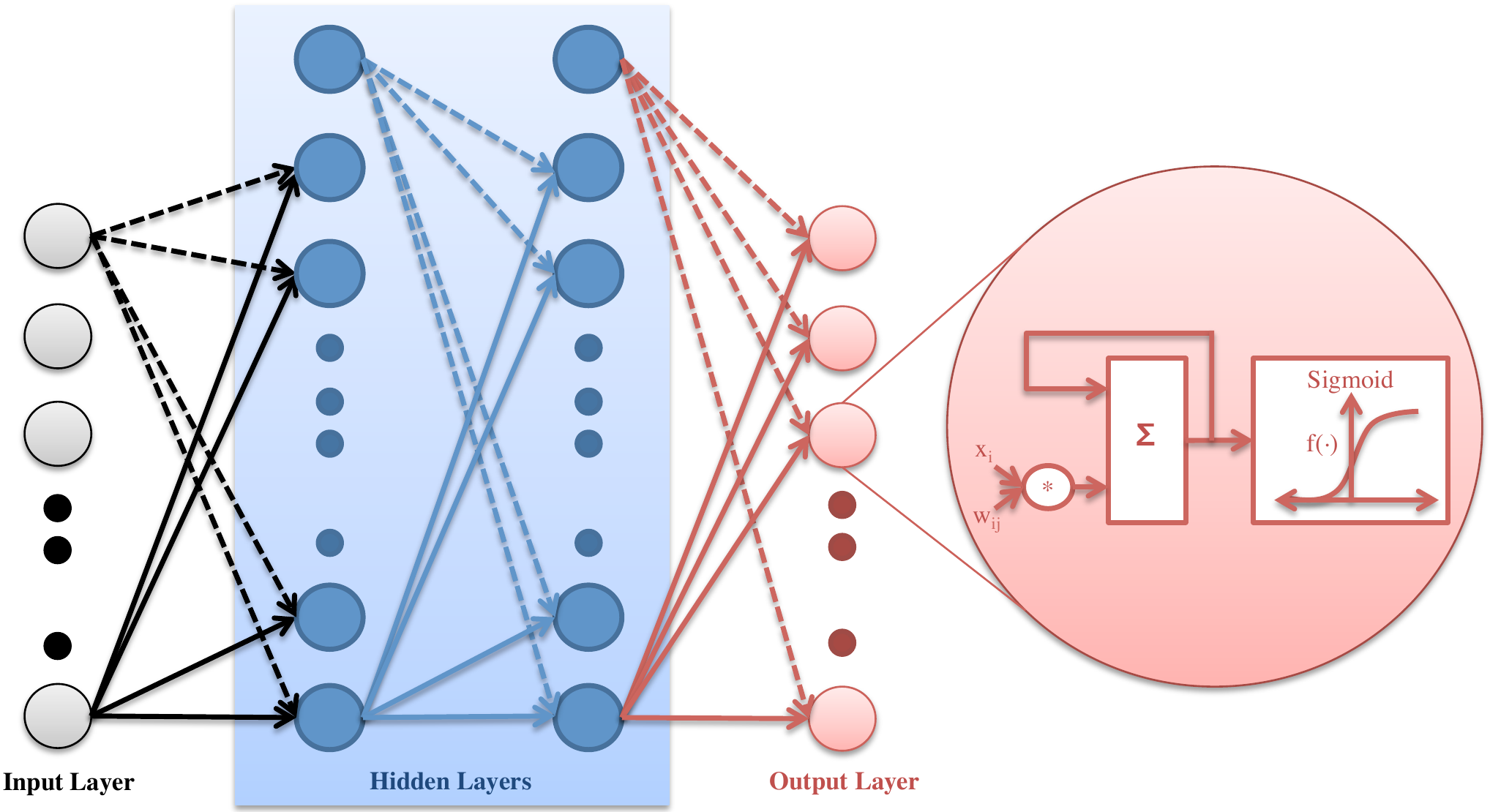}
\caption{Feedforward ANN with an input and two hidden layers, followed by an output layer. The artificial neurons accumulate the product of the inputs and the interconnecting synaptic weights, and apply a sigmoid activation function to the resulting sum.}
\label{fig1}
\end{figure}

In our work, we focus on multilayered ANNs, which possess intrinsic error resiliency [7]. This can be attributed to the inherent redundancy in feature representation, which stems from the observation that information is stored in a distributed manner among a group of neurons. Hence information is not completely lost due to errors introduced in a few of the neurons, or modest perturbations in the stored synaptic weights. Significant energy benefits have been demonstrated in previous literature, using approximate neural processing elements, and scaling the precision of the inputs and synaptic weights while paying a negligible performance penalty [8].

We note that the number of synapses is orders of magnitude larger than the number of neurons. Hence, the on-chip synaptic memory contributes substantially to the power consumption of a typical digital CMOS implementation of ANNs. We propose scaling the supply voltage of the system in order to achieve energy efficiency. The digital logic comprising the neural processing elements and the associated controllers could be operated reliably at scaled voltages by clocking them at a lower frequency. However, the 6T SRAM that has been the workhorse of on-chip memories is prone to bitcell failures at scaled voltages. The process parameter variations could have a detrimental impact on the relative strength of the transistors, resulting in an imbalance in the bitcell that was engineered to have symmetric operation [9]. Supply voltage scaling exacerbates the impact of parameter variations, which could potentially cause a bitcell to experience read access, read disturb, or write failures [10].

We analyzed the impact of voltage scaling on the performance of ANNs on MNIST [11], which is a widely used handwritten digit recognition dataset. Our analysis indicates that the inherent error resiliency of ANNs enables the supply voltage to be moderately scaled for a negligible degradation in the classification accuracy. However, aggressive scaling causes a substantial deterioration in the accuracy due to increased probability of failures in the MSBs of the synaptic weights.

We, therefore propose a significance driven hybrid 8T-6T SRAM, wherein the sensitive MSBs of the synaptic weights are stored in 8T bitcells while the relatively resilient LSBs are stored in 6T bitcells. The 8T bitcells provide stable operation under scaled supply voltage conditions, primarily due to the presence of independently optimized read and write paths. This enables aggressive voltage scaling of the hybrid array for a minimal degradation in the classification accuracy, albeit with an area penalty.

The area overhead can be further minimized by judiciously selecting the number of sensitive MSBs for the synaptic weights interconnecting different layers of an ANN. We propose a synaptic-sensitivity driven hybrid memory architecture consisting of multiple 8T-6T SRAM banks, each of which stores synapses carrying a definite significance. This is motivated by the intuition that the synapses fanning out of a layer of significant neurons need to be protected from large perturbations, over those fanning out of resilient neurons.  It provides improvements in memory access power at reduced area costs. 
The key contributions of our work are:
\begin{enumerate}
\item We show that the supply voltage of a typical digital CMOS implementation of ANNs can be scaled to achieve energy efficiency; despite the susceptibility of a 6T SRAM based synaptic storage to bitcell failures. 
\item In an effort to scale the voltage even further, we propose a significance driven hybrid 8T-6T SRAM that stores the sensitive MSBs in robust 8T bitcells.
\item We further present a synaptic-sensitivity driven hybrid memory architecture, wherein the number of MSBs of the synaptic weights that need to be stored in 8T bitcells are chosen based on their sensitivity so as to gain power benefits with minimal area overheads.
\end{enumerate}
  
The rest of the paper is organized as follows. Section II provides a brief introduction on ANNs. Section III describes the proposed memory architecture and the significance driven synaptic memory design. Section IV analyzes the bitcell failures and power consumption of both 6T and 8T SRAM arrays at scaled supply voltages. Section V explains the simulation methodology. Section VI presents the results while Section VII concludes the paper.
\begin{figure}[!t]
\centering
\includegraphics[width=2.8in, height =2in]{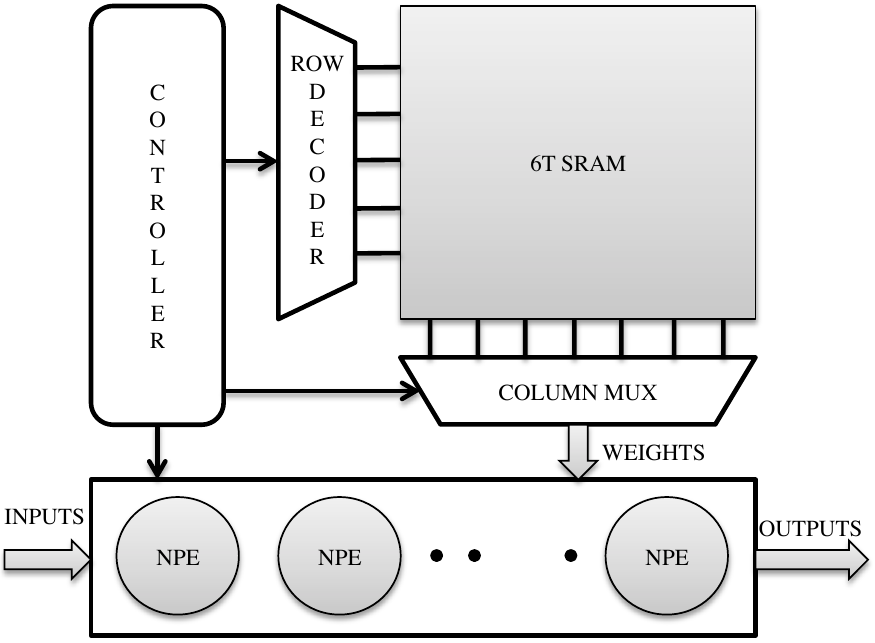}
\caption{Block diagram of a digital ASIC implementation of ANNs. Neural Processing Elements (NPEs) mimic the computations of the artificial neurons, and an on-chip 6T SRAM is used for storing the synaptic weights.}
\label{fig2}
\end{figure}
\section{Artificial Neural Networks}
ANNs are composed of layers of artificial neurons interconnected by synapses, and approximately mimic certain types of computations performed by the human brain. The neurons are arranged in different layers \textit{viz.} input, one or more hidden, and output layers. Every layer in the ANN saving the output layer is fully connected to the layer immediately following it, and such a network with acyclic synaptic connectivity from the input to the output layer is known as a feedforward ANN. Every neuron in the ANN with the exception of input neurons sums the product of the incoming inputs and connecting weights. Subsequently, a nonlinear function, for instance, the sigmoid function is applied to the summed result in order to obtain the activation output.

An ANN is trained in a \textit{supervised} manner using the backpropagation algorithm [12] on a designated training dataset. The key idea is to iteratively update the synaptic weights in a manner that minimizes their contribution to the output error. A trained neural network stores the learned features in a distributed manner. It is this property that enables an ANN to tolerate errors in neuronal computations and synaptic weight perturbations. We exploit the error resiliency to architect a power-efficient on-chip synaptic memory as will be described in the following section.

\section{Synaptic Memory Architecture}
This section describes the proposed synaptic memory designs for a typical neuromorphic system shown in Fig. 2. It consists of Neural Processing Elements (NPEs) that mimic the core computations of the artificial neurons, and a conventional 6T SRAM for on-chip synaptic storage. It additionally requires a controller to coordinate the sequence of operations between the NPEs and the synaptic memory.

Digital CMOS implementations of ANNs are innately power hungry due to the heavy computational demands placed on the NPEs, and the synaptic memory access and leakage power consumption. Supply voltage scaling can be used to achieve significant energy savings. The NPEs and the associated control logic can be reliably operated at scaled voltages by lowering the clock frequency. However, a 6T SRAM is susceptible to bitcell failures under scaled voltage conditions. The failures are aggravated by process parameter variations in scaled technology nodes. On the other hand, ANNs being error resilient applications can tolerate moderate perturbations in their constituent synaptic weights. Nevertheless, the classification accuracy might deteriorate substantially if the MSBs of a large fraction of the synapses are corrupted. Hence, the stability of bitcells that store the MSBs is of paramount importance for aggressive voltage scaling. This intuition is central to the significance and sensitivity based memory designs that will be described subsequently.

\subsection{Configuration 1: Significance Driven Hybrid 8T-6T SRAM}
In a significance driven hybrid array, few MSBs of all the synaptic weights are stored in 8T bitcells as illustrated in Fig. 3(b). An 8T bitcell has read and write paths that can be independently optimized for the respective operations as opposed to a shared path in 6T bitcells. This enables the voltage of the hybrid array to be scaled aggressively. This leads to improvements in memory access and leakage power consumption, however at the cost of an increase in area. We use additional application-level insights, and present an improved memory architecture to minimize the area penalty, as described in the following sub-section.

\begin{figure}[!t]
\centering
\includegraphics[width=2.8in, height = 3.5in]{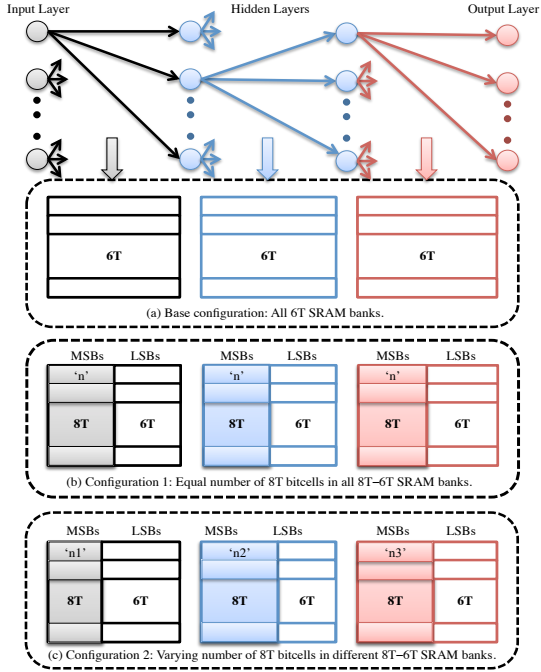}
\caption{Synaptic memory configurations under study. (a) All 6T SRAM. (b) Significance driven hybrid 8T-6T SRAM. (c) Synaptic-sensitivity driven hybrid memory architecture.}
\label{fig3}
\end{figure}

\subsection{Configuration 2: Synaptic-Sensitivity Driven Hybrid Memory Architecture}
The basic significance driven architecture protected an equal number of MSBs in all the synaptic weights. We note that additional power benefits can be obtained with minimal area overheads by reducing the number of 8T bitcells. Fig. 3(c) shows a block diagram of the proposed synaptic-sensitivity driven hybrid memory architecture. It consists of multiple 8T-6T SRAM banks, each of which stores the synapses fanning out of neurons in the corresponding layer of an ANN. The number of MSBs of the synaptic weights that needs to be stored in 8T bitcells is chosen based on their sensitivity.

In a deep ANN, the first hidden layer extracts the low level features from the input dataset. Furthermore, a reasonable fraction of the synapses are concentrated in the input and the initial hidden layers, since the number of neurons per layer decreases progressively from the input to the output layer [14]. In general, large synaptic weight perturbations in the input and the first hidden layer could have a detrimental effect on the classification accuracy, and are hence deemed significant. It has been shown that the fraction of resilient neurons decreases while moving towards the output layer [8]. Intuitively, injecting substantial errors in the synapses fanning into the output layer would directly impact the classifier performance. Hence the output layer is more sensitive than the central hidden layers. The resilient synapses have relatively fewer MSBs stored in 8T bitcells in comparison to the ones that were deemed significant. The proposed memory architecture thus exploits the varying significance of the synapses connecting different layers in order to provide power savings at reduced area costs.

A rigorous analysis of 8T and 6T bitcells is required at the circuit level in order to quantify the power benefits and area overhead associated with each of the configurations described in this section and shown in Fig. 3. The following section presents a comprehensive failure analysis of both the 8T and 6T bitcell topologies.

\section{Failure Analysis of 6T and 8T SRAMS}
	
Failures in SRAMs arise due to random variations in the process parameters [15–--17]. Among the different sources of random intra-die variations, the most prominent one is the threshold voltage variation that is caused by random dopant fluctuations [10]. We have therefore considered only the failures caused due to on-die variations in the threshold voltage without loss of generality.

The different types of failures in an SRAM are:
\begin{enumerate}
\item Read access failure, which is caused by an inability to complete a successful read operation before the end of the read cycle.
\item Write failure, which is caused by an inability to successfully flip a bit within the stipulated write time.
\item Read disturb failure, which is caused by unintended flipping of the stored data during a read operation.
\end{enumerate}

We designed a 6T SRAM bitcell shown in Fig. 4(a) in 22 nm technology using predictive models [18]. It is sized to have a nominal static read noise margin of 195 mV, which is a measure of the robustness of an SRAM bitcell against flipping. The write margin, which is a measure of the easiness to write into an SRAM bitcell is 250 mV. A 6T SRAM has conflicting read and write sizing requirements [10] making it susceptible to failures at scaled voltages. On the other hand, an 8T SRAM shown in Fig. 4(b) can be optimized separately for read and write operations. The 6T and 8T bitcells were designed for equal read access and write times, which were determined by considering the delay incurred in charging/discharging the bitline capacitance associated with a 256x256 SRAM sub-array. 

The SRAM bitcells were then subjected to threshold voltage ($V_T$)  fluctuations. The $V_T$ fluctuations ($\Delta V_T$) in the transistors in an SRAM bitcell are considered as independent Gaussian random variables with zero mean [19]. The standard deviation of the  $V_T$ fluctuations ($\sigma_{V_T}$) is a strong function of the transistor sizes [10, 20], and the dependency is given by

\begin{equation}
\sigma_{V_T} = \sigma_{V_{T0}} \sqrt{\left(\frac{L_{min}}{L}\right)\left(\frac{W_{min}}{W}\right)}
\end{equation}
Where $\sigma_{V_{T0}}$ is the standard deviation of a minimum sized transistor, $L_{min}$ and $W_{min}$ are the minimum allowed length ($L$) and width ($W$) of the technology node respectively.

\begin{figure}[!t]
\centering
\includegraphics[width=3.2in]{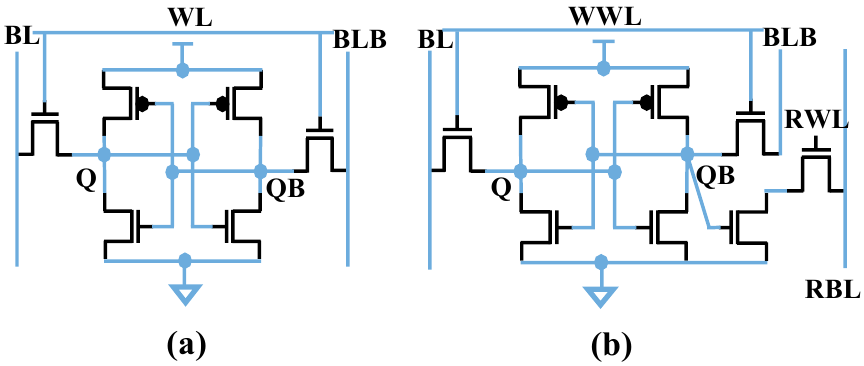}
\caption{(a) 6T SRAM bitcell. (b) 8T SRAM bitcell. The 8T bitcell consists of two additional transistors for decoupled read and write operations.}
\label{fig4}
\end{figure}

Finally, Monte Carlo simulations were run on a 256x256 SRAM sub-array to estimate the read access, read disturb, and write failure rates at different operating voltages. It can be seen from Fig. 5 that read access failures dominate over write failures in a 6T SRAM at scaled voltages. The corresponding failures for an 8T SRAM are negligible in the voltage range of interest. Similarly, we found that the read disturb failures were small enough to be neglected for a 6T SRAM while an 8T SRAM is free from disturb failures [21]. 
	
Fig. 6 shows the variation of memory access and leakage power with supply voltage scaling. It can be seen that an 8T bitcell consumes roughly 20\% more read and write power, and 47\% more leakage power than a 6T bitcell under iso-voltage conditions. Our layout analysis indicates that the 8T bitcell incurs a 37\% area overhead. We further note that the hybrid 8T-6T arrays can effectively be laid out in a single row [13], and hence does not incur any other overhead aside from the obvious area and power penalty owing to an increase in the transistor count. The 8T and 6T bitcell characteristics thus obtained would be used to evaluate the proposed synaptic memory designs. The simulation methodology is described in the following section.

\begin{figure}[!t]
\centering
    \subfloat[]{{\includegraphics[width=1.55in]{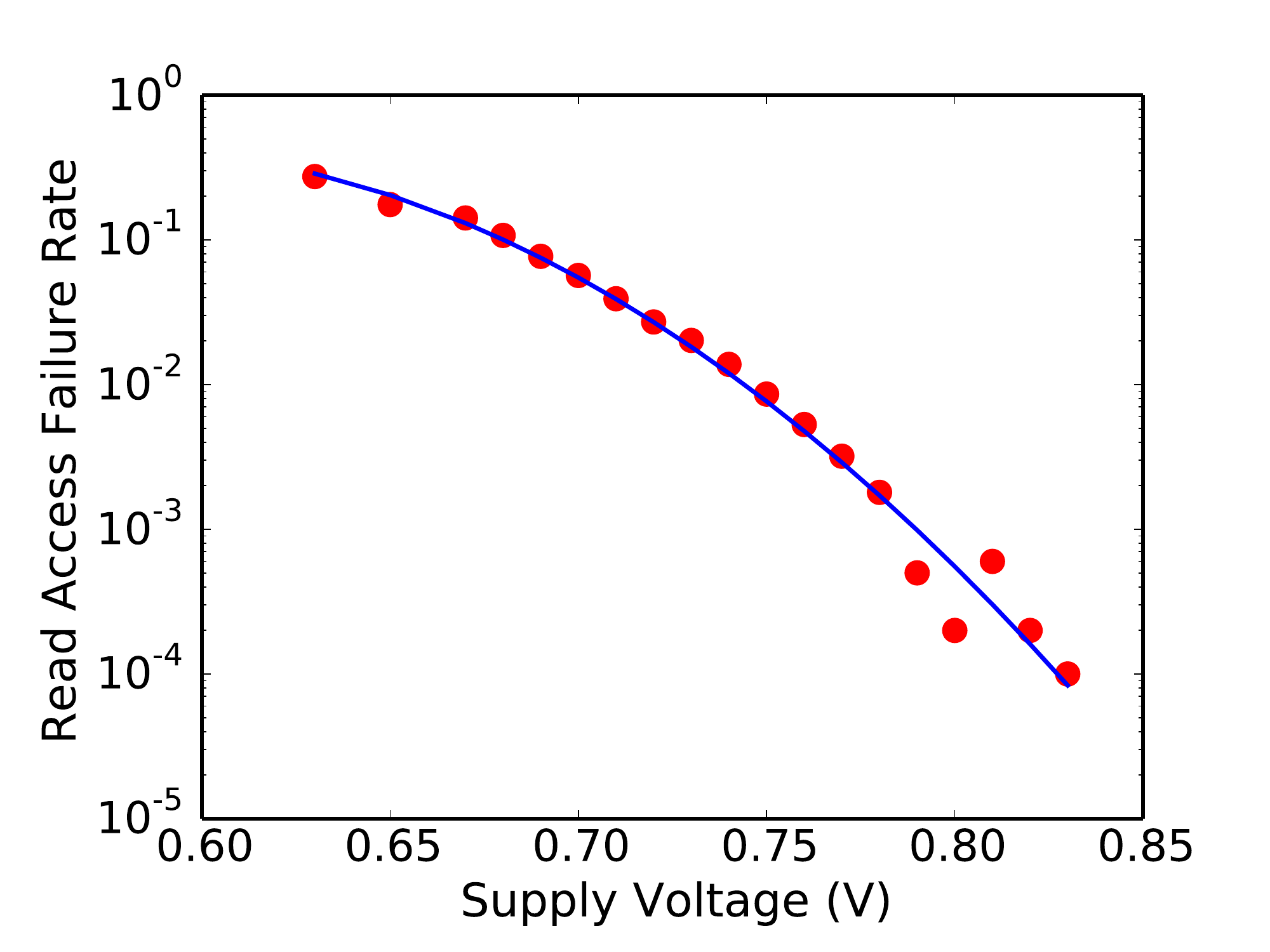} }}%
    \qquad
    \subfloat[]{{\includegraphics[width=1.55in]{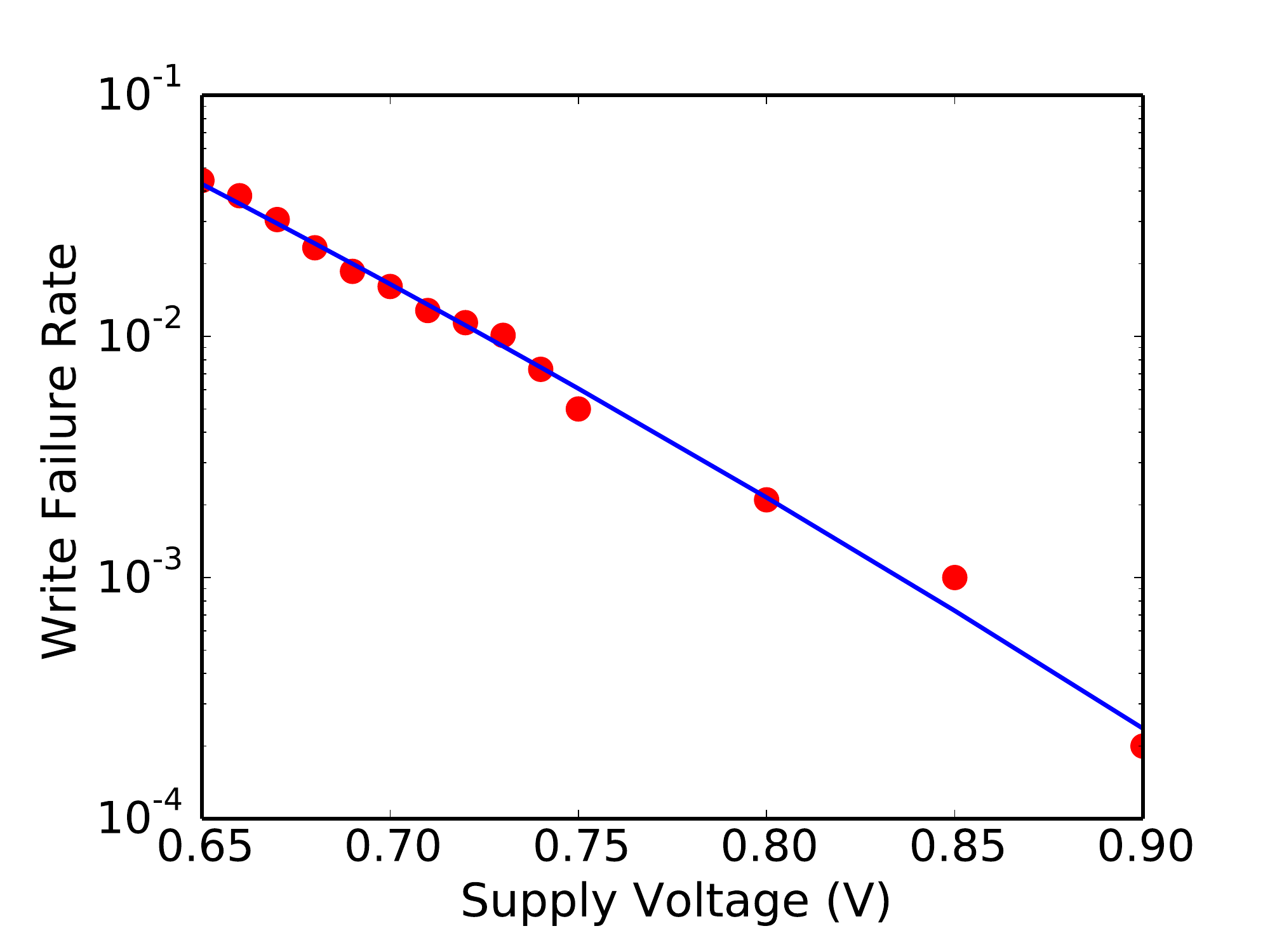} }}%
\caption{(a) Read access failure rate versus supply voltage for a 6T SRAM bitcell. (b) Write failure rate versus supply voltage for a 6T SRAM bitcell.}%
\end{figure}

\begin{figure*}
    \centering
    \subfloat[]{{\includegraphics[width=5cm]{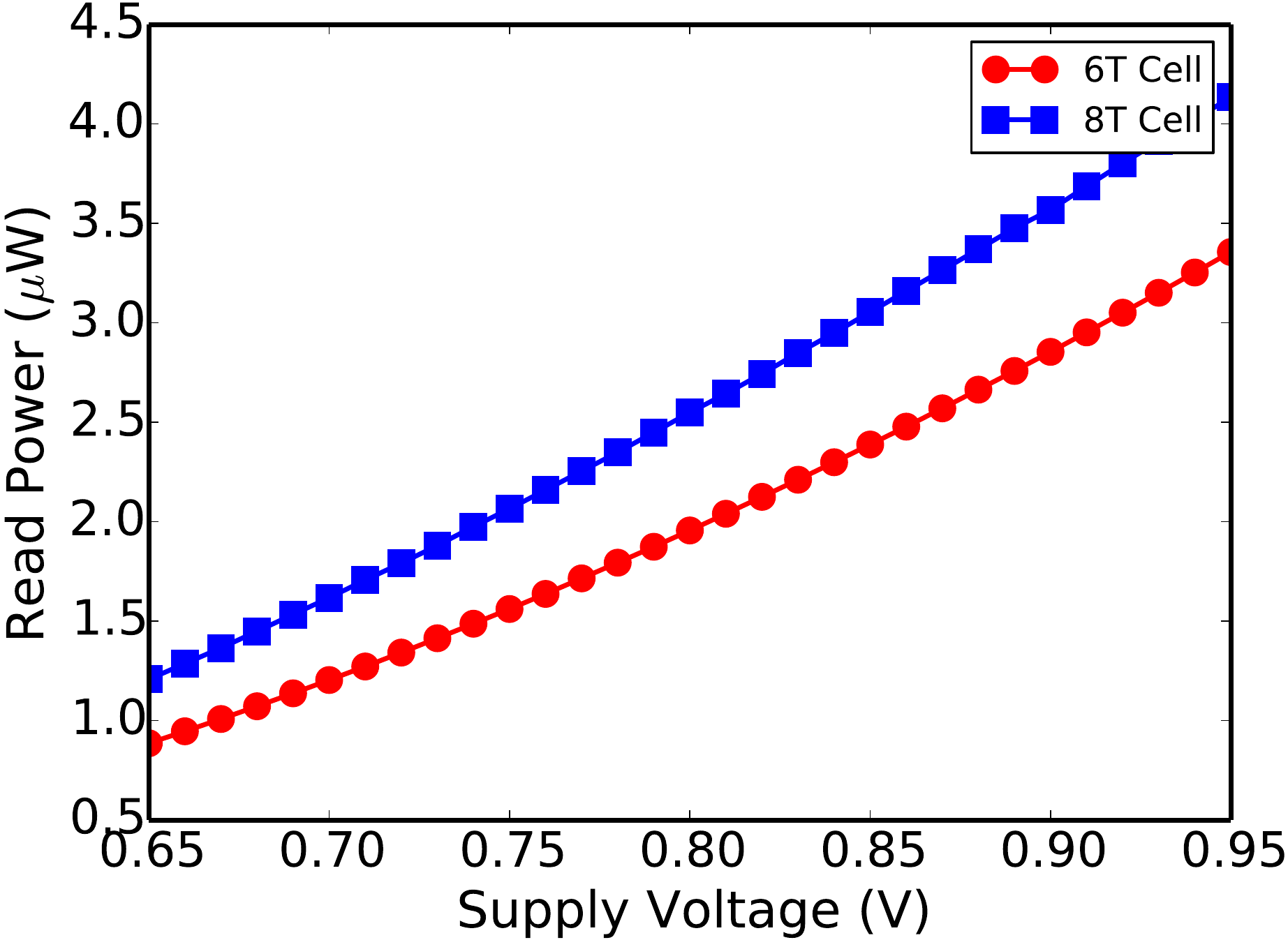} }}%
    \qquad
    \subfloat[]{{\includegraphics[width=5cm]{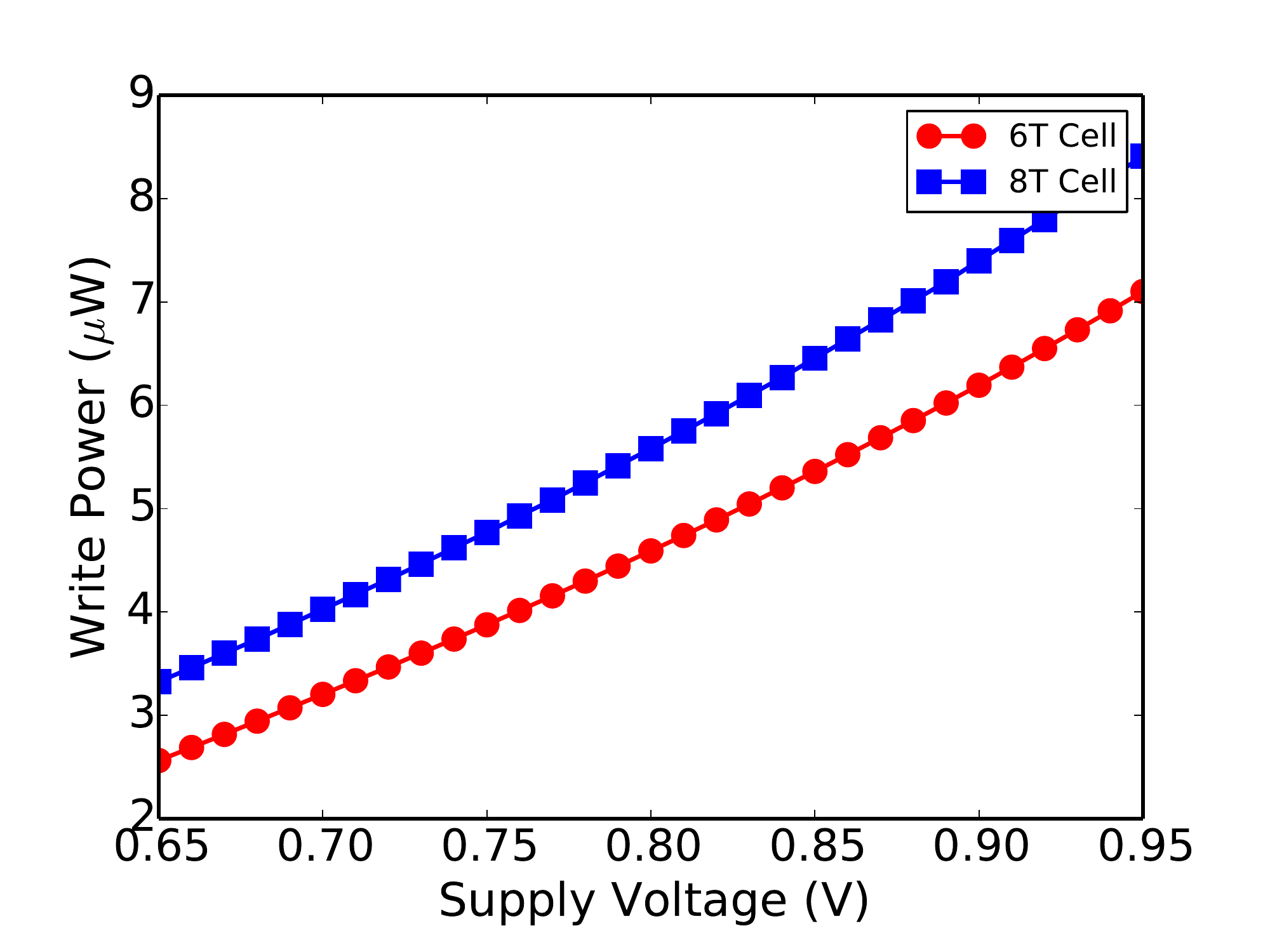} }}%
    \subfloat[]{{\includegraphics[width=5cm]{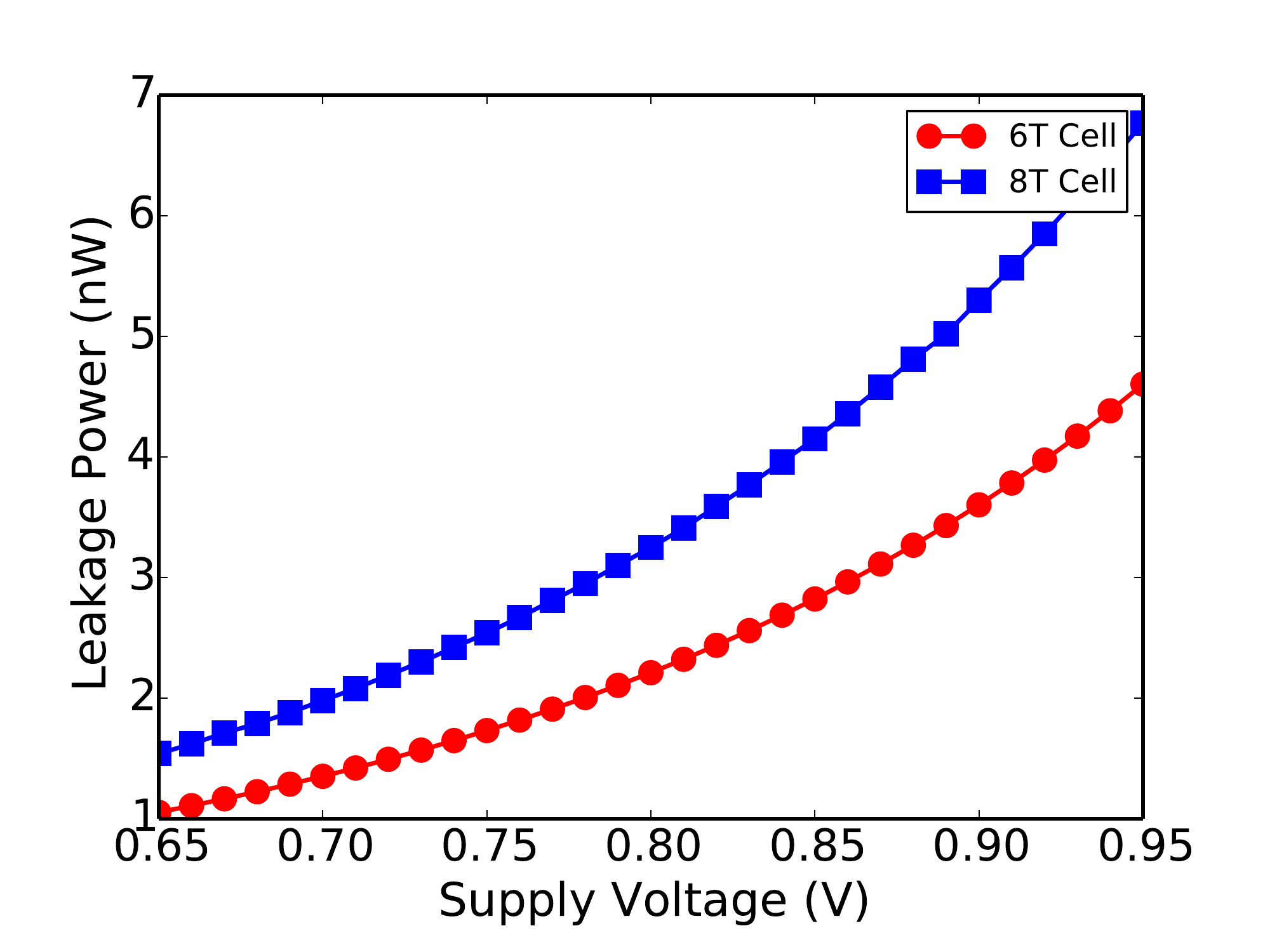} }}%
    \qquad
    \caption{(a) Read power versus supply voltage, (b) write power versus supply voltage, and (c) leakage power versus supply voltage of 6T and 8T bitcells.}%
    \label{fig:example}%
\end{figure*}

\section{Simulation Methodology}
A circuit to system-level simulation framework was developed to analyze the impact of voltage scaling on the proposed synaptic memory designs. At the circuit level, the 6T and 8T bitcells were designed, and subjected to SPICE simulations to estimate the area, power, and failure rates. The failure probabilities and the different synaptic memory configurations \textit{viz.} 6T SRAM, hybrid 8T-6T SRAM are fed to an ANN functional simulator.

At the system level, the deep learning toolbox [22], which is an open source neural network simulator, was used to train and evaluate the performance of the ANN under consideration. The read access and write failures are modeled by introducing bit flips while accessing and updating the synaptic weights in the functional simulator. The distribution of bit failures depends on the synaptic memory configuration, for instance, the failures are distributed uniformly for a 6T SRAM while only the LSBs are affected in a hybrid 8T-6T SRAM. The failure analysis of 8T bitcell conclusively proves that it is virtually unaffected by supply scaling within the voltage range of interest. It was additionally assumed that a 6T bitcell cannot simultaneously have read access and write failures since they necessitate conflicting requirements. The simulator computes the degradation in classification accuracy owing to synaptic weight perturbations. The bitcell characteristics determined from SPICE simulations together with the synaptic memory configuration are used to calculate the area, memory access and leakage power consumption.

\subsection{ANN Benchmark}
The benefits of the proposed synaptic memory designs are evaluated on a multilayered feedforward ANN that is trained to classify handwritten digits. The essential parameters of the benchmark ANN are shown in Table I.

\begin{table}[h]
\renewcommand{\arraystretch}{1.1}
\centering
\caption{ANN Architecture For Digit Recognition}
\label{table5}
\resizebox{\columnwidth}{!}{%
\begin{tabularew}{lcccccc}

\hline
\hline
Data Set    & Num. Layers    & Num. Neurons   & Num. Synapses     \\ 
MNIST   &    6        & 2594    & 1406810     \\ 

 \hline \hline

\end{tabularew}
}

\end{table}

\section{Results and Discussions}
In this section, we present the results that demonstrate the trade-offs between classification accuracy, power and area for the proposed synaptic memory designs. As noted earlier, we primarily focus on the on-chip synaptic memory storage, since the digital neurons and the required control logic can be operated reliably at scaled voltages. We use a synaptic precision of 8 bits since the observed degradation in accuracy is less than 0.5\% from the nominal value, which corresponds to a precision of 32 bits.

\subsection{Performance and power trade-offs of 6T SRAM}
Fig. 7(a) shows the impact of voltage scaling on the classification accuracy of an ANN, when a 6T SRAM is used for on-chip synaptic weight storage. The results indicate that the intrinsic error resiliency of the ANN allows the voltage to be scaled up to 200 mV from the nominal operating voltage (950 mV) for almost no loss (less than 0.5\%) in the classification accuracy. Fig. 7(b) illustrates the savings in memory access and leakage power as a consequence of supply scaling. Our analysis further indicates that aggressive scaling results in a degradation of more than 30\% in the accuracy owing to substantial errors in the MSBs of the synaptic weights.

\subsection{Performance and power trade-offs of the significance driven hybrid 8T-6T SRAM}
Fig. 8 illustrates that a hybrid 8T-6T SRAM, wherein a few MSBs of all the synaptic weights are stored in 8T bitcells allows the voltage to be scaled by another 100 mV. The aggressive voltage scaling is made possible because of the robust operation of 8T bitcells at reduced voltages. An iso-stability analysis was carried out to demonstrate the power benefits. A 6T SRAM operating at 0.75 V was used as the baseline synaptic memory configuration in order to evaluate the improvement in power consumption, and the corresponding area overheads. The results show that protecting three MSBs of all the synapses provides a 29\% improvement in memory access and leakage power consumption, albeit at a 13.75\% area penalty. The error resiliency of the ANN to LSBs of the synaptic weights precludes the need for an all 8T SRAM. Fig. 8(a) conclusively proves that protecting three or four MSBs in 8T bitcells is sufficient to achieve close to nominal accuracy.

\begin{figure*}
\centering
    \subfloat[]{{\includegraphics[width=2.5in]{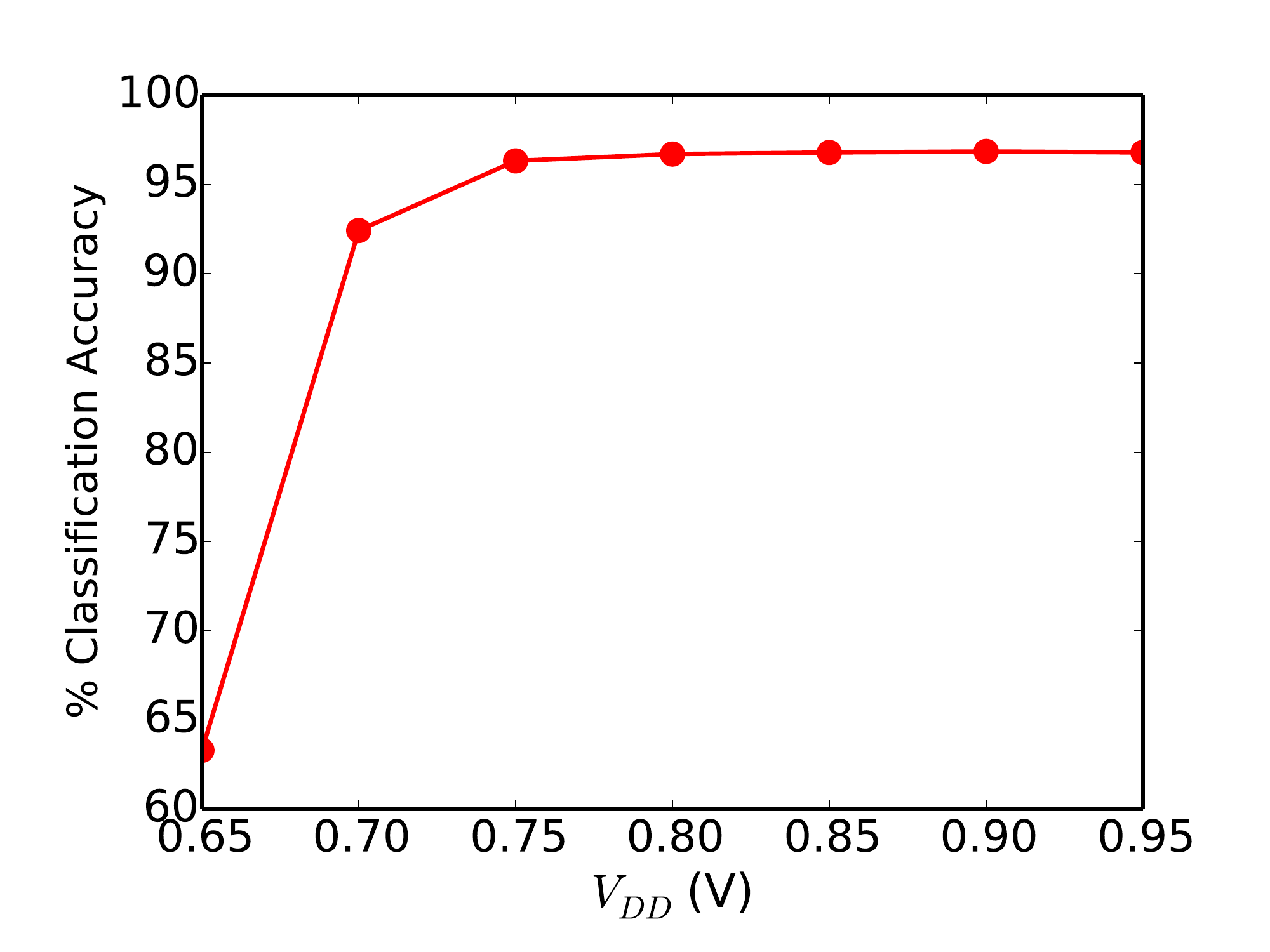} }}%
    \qquad
    \subfloat[]{{\includegraphics[width=2.5in]{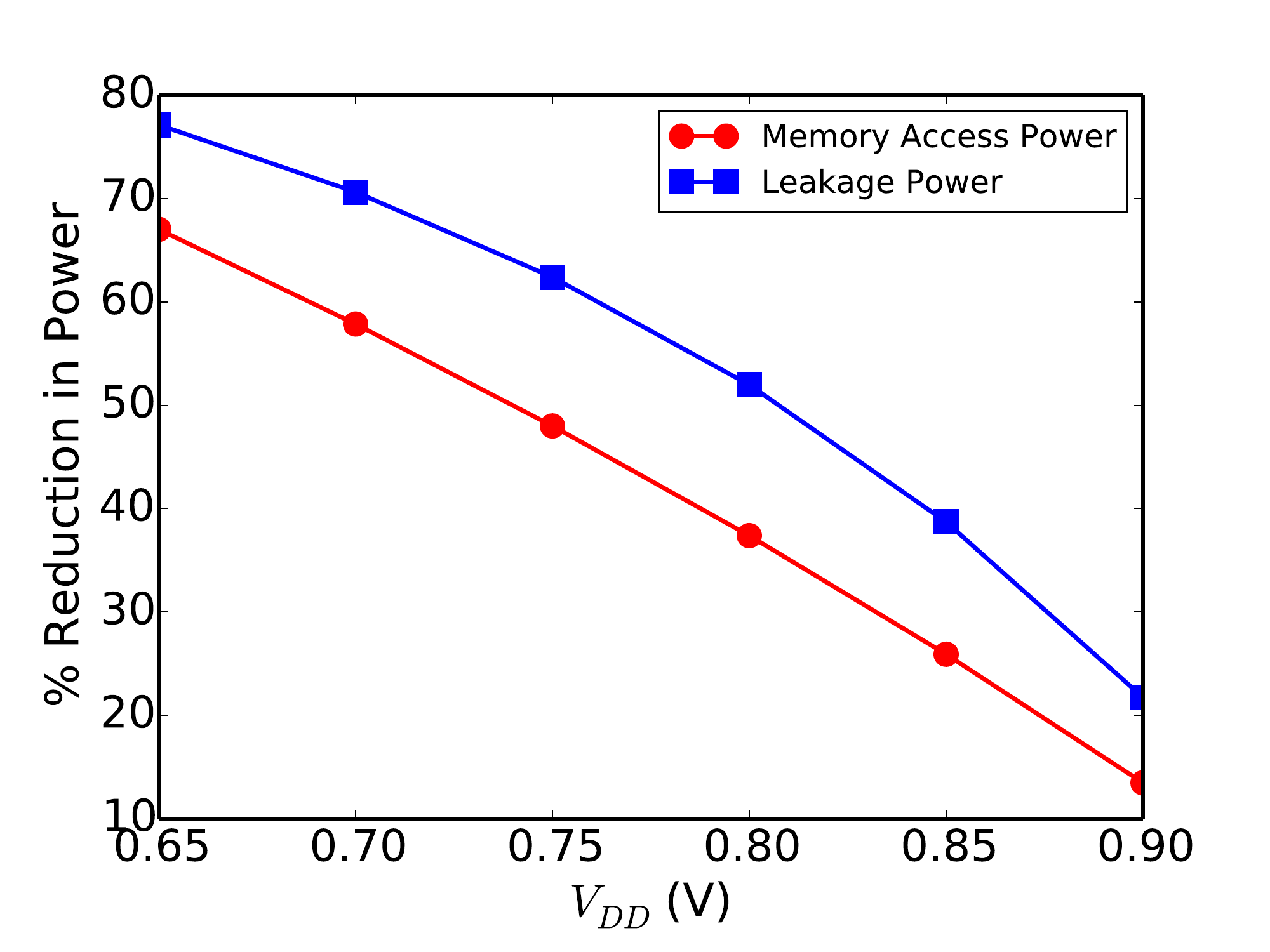} }}%

 \caption{(a) Classification accuracy versus $V_{DD}$ for 6T SRAM. (b) Memory access and leakage power savings versus $V_{DD}$ for 6T SRAM.}%
\end{figure*}

\begin{figure*}

    \centering
    \subfloat[]{{\includegraphics[width=5cm]{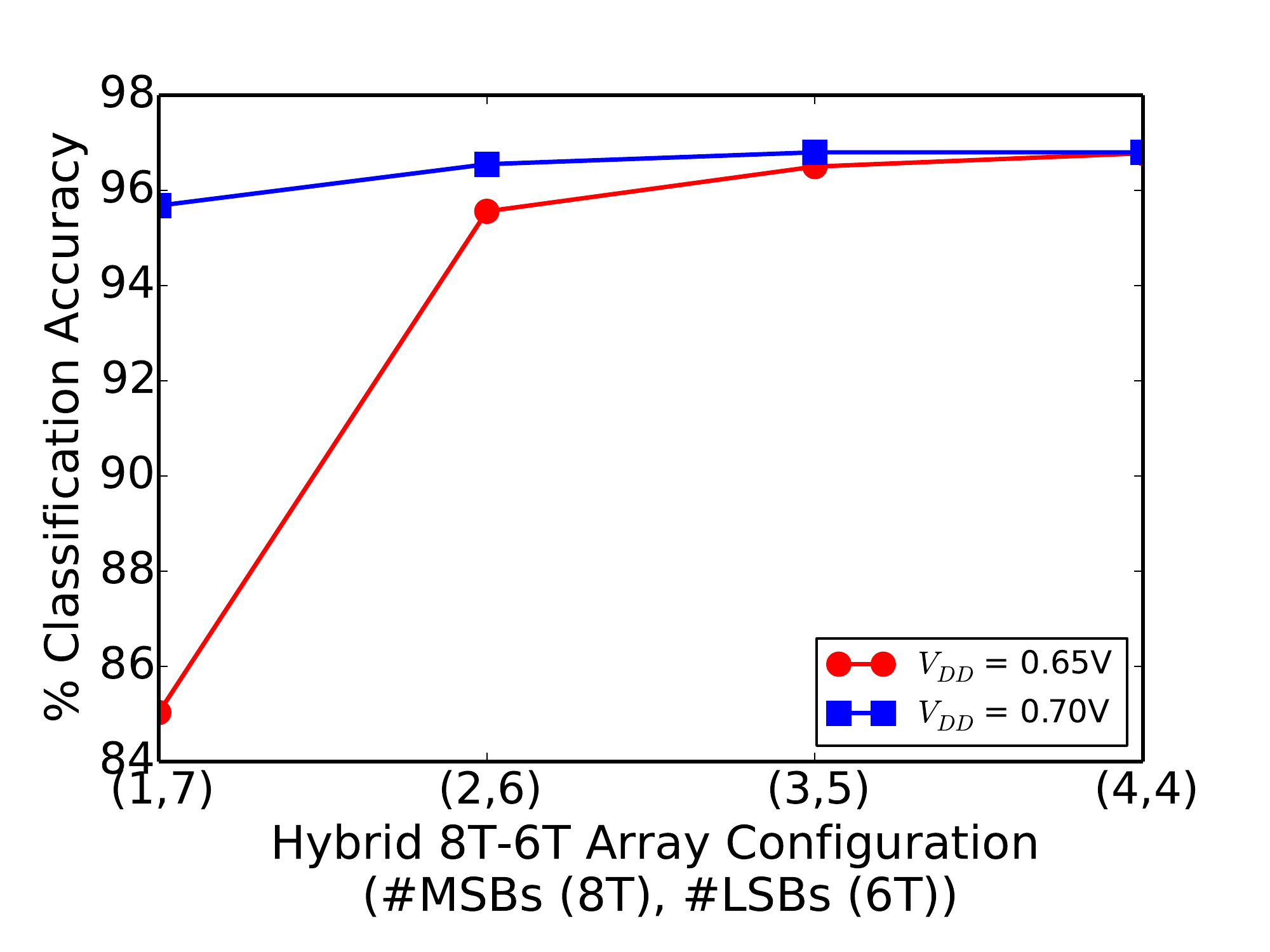} }}%
    \qquad
    \subfloat[]{{\includegraphics[width=5cm]{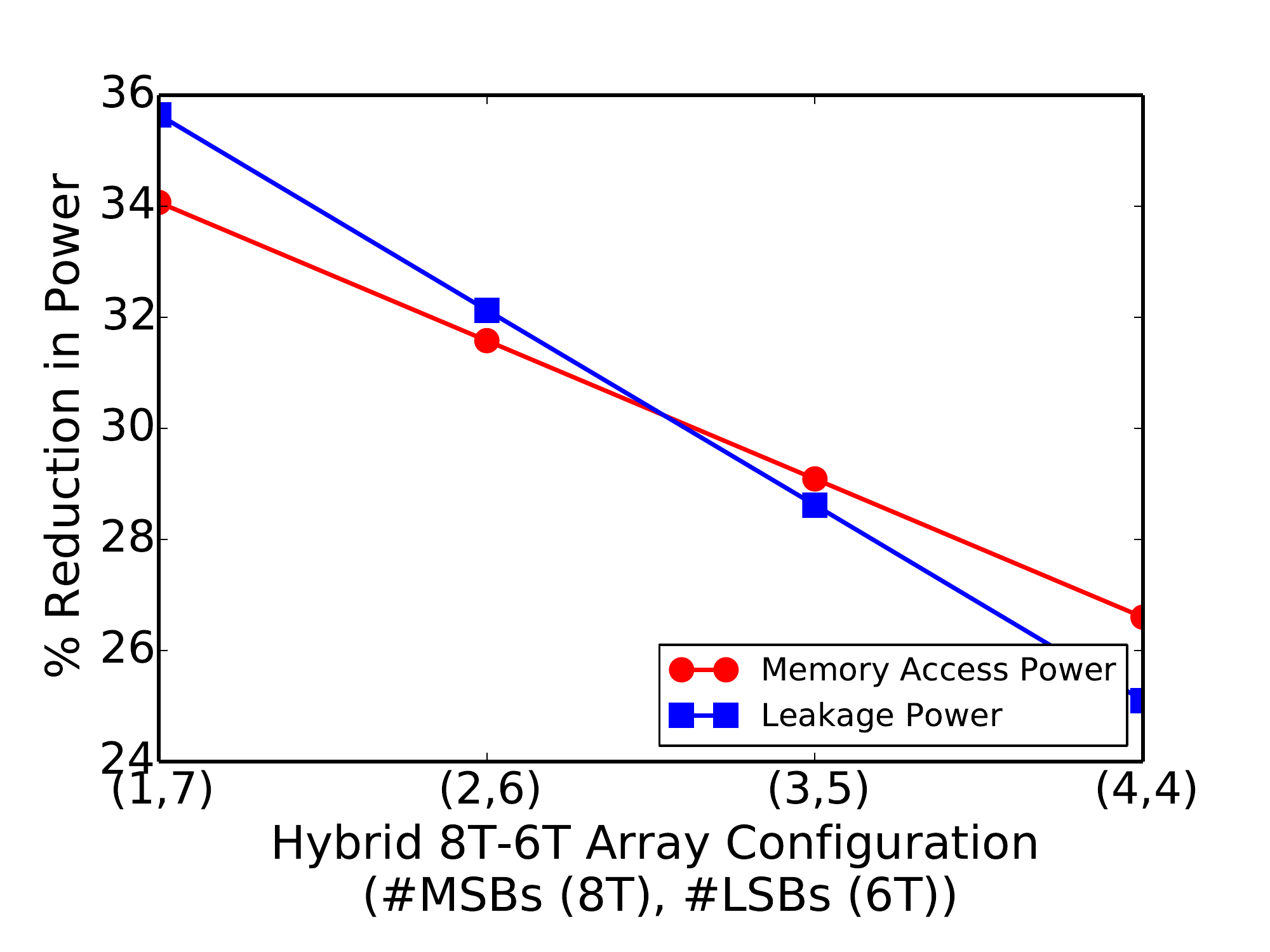} }}%
    \subfloat[]{{\includegraphics[width=5cm]{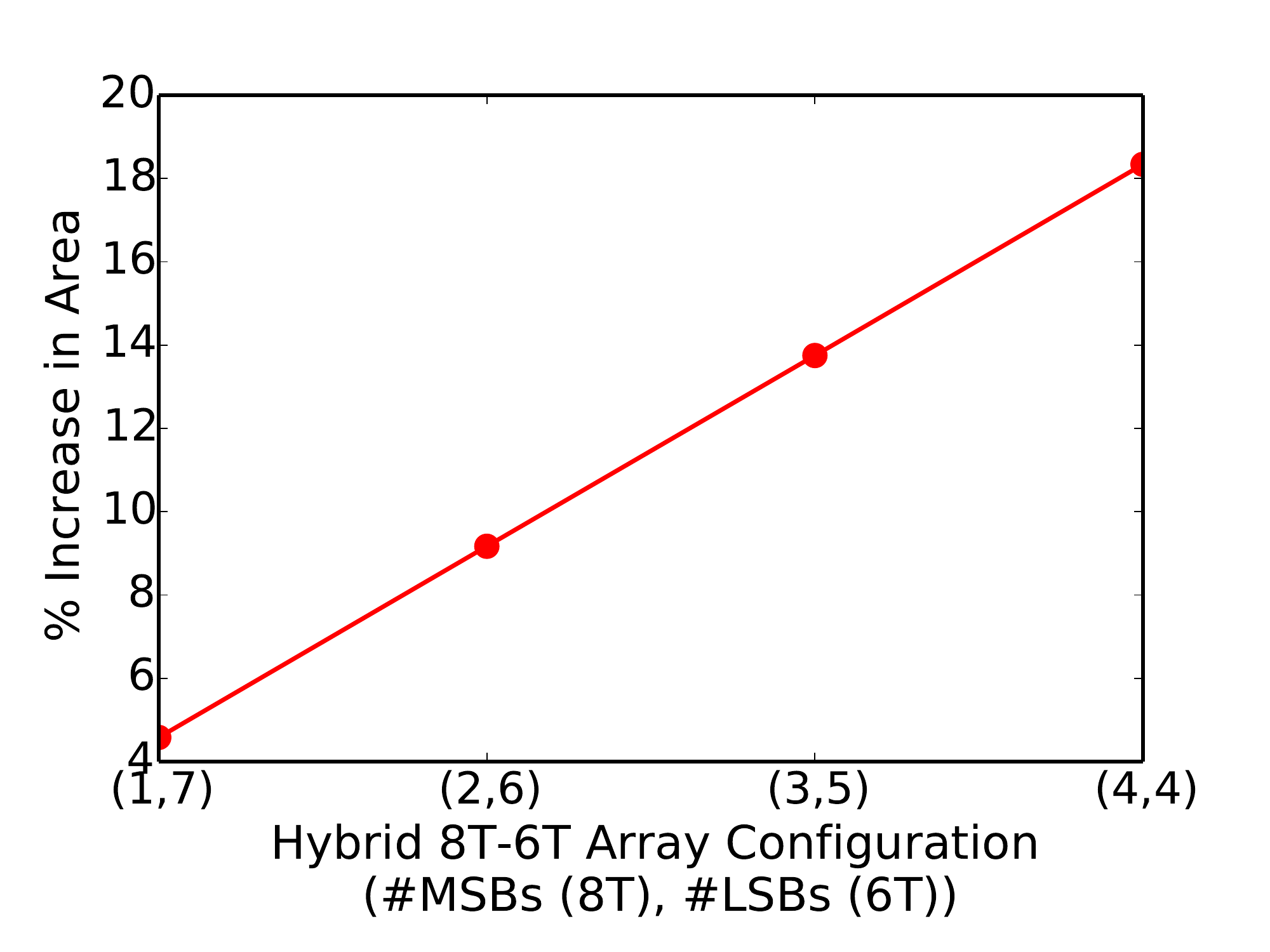} }}%
    \qquad
    \caption{(a) Classification accuracy, (b) memory access and leakage power benefits ($V_{DD}$ = 0.65 V), and (c) area overhead for various 8T-6T array configurations.}%
    \label{fig:example}%

\end{figure*}

\subsection{Performance and power trade-offs of the synaptic-sensitivity driven hybrid memory architecture}
This memory architecture consists of five 8T-6T SRAM banks, each of which store the synapses fanning out of neurons in the corresponding layer of the benchmark ANN. The following intuitions were used to determine the sensitivity of synapses, which are corroborated by the results shown in Fig. 9.
\begin{enumerate}
\item The synapses fanning out of the input and first hidden layer are significant in comparison to those interconnecting the central hidden layers.
\item The synapses fanning into the output layer are important, since any errors directly impact the classifier output.
\end{enumerate}

Our analysis further indicates that the input layer is resilient relative to the first hidden layer. Intuitively, this can be attributed to the fact that the input images typically contain a slew of insignificant pixels along with the features of interest. For instance, the digits in the MINIST dataset are concentrated in the center. Thus, the pixels at the image boundaries do not contain useful information.  The ability of the input layer to tolerate synaptic errors better than the first hidden layer yields power savings while further reducing the area costs. The results show that a 30.91\% reduction in memory access power can be obtained with a 10.41\% area overhead for less than 1\% loss in the classification accuracy. Additional power savings of 7.38\% could be achieved at a further 40.25\% reduction in the area cost, if a degradation of less than 4\% in the classification accuracy could be tolerated.

\section{Conclusion}
In this work, we explored scaling the supply voltage of large-scale neuromorphic systems to achieve energy efficiency. In scaled technologies, under supply voltage scaling, an on-chip synaptic memory designed using a standard 6T SRAM is susceptible to bitcell failures. We took advantage of the ability of ANNs to tolerate modest errors in the synaptic weights to minimize the power consumption. Nevertheless, aggressive scaling resulted in substantial performance degradation due to errors in MSBs of the synaptic weights. To this end, we proposed a significance driven hybrid 8T-6T SRAM, wherein a few MSBs of the synaptic weights are stored in robust 8T bitcells. The hybrid array yielded substantial power savings, since it allowed the voltage to be scaled lower than that which could be achieved using a 6T SRAM. We finally presented a synaptic-sensitivity driven hybrid memory architecture. We availed the varying significance of synapses connecting different layers of the ANN to gain power benefits with minimal area overheads. Thus, we harnessed the significance driven computing methodology and error resiliency of ANNs to architect an efficient on-chip synaptic storage.

\begin{figure}[!t]
\centering
\includegraphics[width=2.3in]{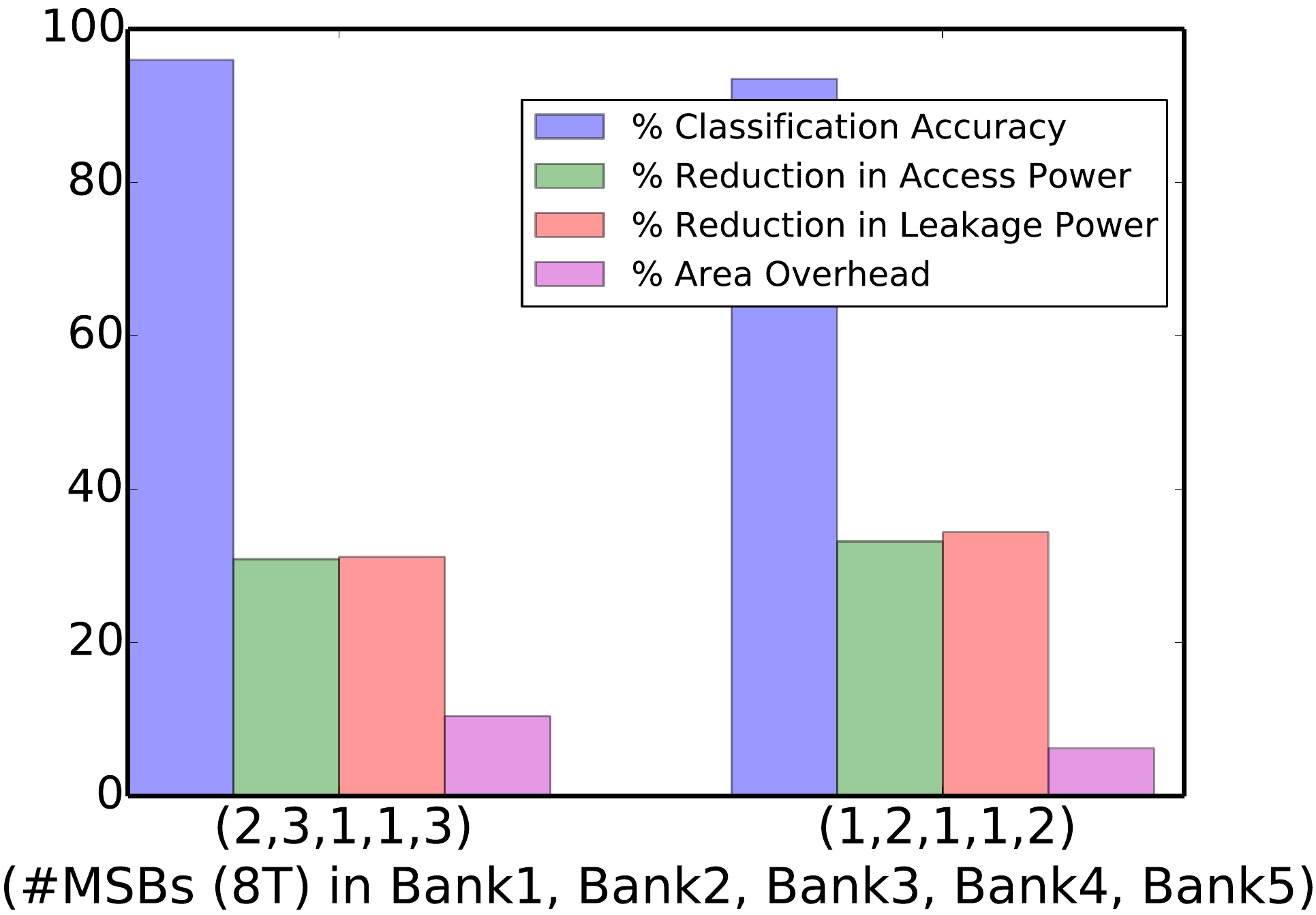}
\caption{Classification accuracy, memory access and leakage power, and area trade-offs for two memory configurations with varying number of 8T bitcells in different 8T-6T SRAM banks operating at a voltage of 0.65 V.} 
\label{fig9}
\end{figure}





%

\end{document}